\documentclass[nohyperref]{article}

\usepackage{microtype}
\usepackage{graphicx}
\usepackage{multirow}
\usepackage{multicol}
\usepackage{graphics}
\usepackage{caption}
\usepackage{subfigure}
\usepackage{booktabs} 
\usepackage{tabularx}
\usepackage{hyperref}

\usepackage[accepted]{icml2022}

\usepackage{amsmath}
\usepackage{amssymb}
\usepackage{mathtools}
\usepackage{amsthm}
\usepackage{physics}
\usepackage{bbm}
\usepackage{amsfonts}       
\usepackage{nicefrac}       
\usepackage{microtype}      
\usepackage{bm}

\newcommand{\angstrom}{\textup{\AA}}
\newcommand*{\R}{\mathbb{R}}

\usepackage[capitalize,noabbrev]{cleveref}

\theoremstyle{plain}

\theoremstyle{definition}

\theoremstyle{remark}

\usepackage[textsize=tiny]{todonotes}

\icmltitlerunning{Equivariant Graph Attention Networks for Molecular Property Prediction}

\begin{document}

\twocolumn[

\icmltitle{Equivariant Graph Attention Networks \\
for Molecular Property Prediction}



\icmlsetsymbol{equal}{*}

\begin{icmlauthorlist}
\icmlauthor{Tuan Le}{{bayer},{fu}}
\icmlauthor{Frank Noé}{fu}
\icmlauthor{Djork-Arné Clevert}{bayer}
\end{icmlauthorlist}

\icmlaffiliation{bayer}{Bayer AG, Machine Learning Research, Berlin, Germany}
\icmlaffiliation{fu}{Freie Universität Berlin, Department of Mathematics and Computer Science, Berlin, Germany}

\icmlcorrespondingauthor{Tuan Le}{tuan.le2@bayer.com}
\icmlcorrespondingauthor{Djork-Arné Clevert}{djork-arne.clevert@bayer.com}

\icmlkeywords{Machine Learning, ICML}

\vskip 0.3in

]



\printAffiliationsAndNotice{}  

\begin{abstract}
Learning and reasoning about 3D molecular structures with varying size is an emerging and important challenge in machine learning and especially in drug discovery. Equivariant Graph Neural Networks (GNNs) can simultaneously leverage the geometric and relational detail of the problem domain and are known to learn expressive representations through the propagation of information between nodes leveraging higher-order representations to faithfully express the geometry of the data, such as directionality in their intermediate layers. In this work, we propose an equivariant GNN that operates with Cartesian coordinates to incorporate directionality and we implement a novel attention mechanism, acting as a content and spatial dependent filter when propagating information between nodes. We demonstrate the efficacy of our architecture on predicting quantum mechanical properties of small molecules and its benefit on problems that concern macromolecular structures such as protein complexes.
\end{abstract}

\section{Introduction}
Predicting molecular properties is of central importance to applications such as drug discovery or protein design. In silico (computational) methods with fast and precise predictions can significantly accelerate the overall process of finding better molecular candidates in a faster and cheaper way.
Learning on $3$D environments of molecular structures is a rapidly growing area of machine learning with promising applications but also domain-specific challenges. 
While Deep Learning (DL) has replaced hand-crafted features to a large extent, many advances are crucially determined through inductive biases in deep neural networks. Developed neural models should maintain an efficient and accurate representation of structures with even up to thousand of atoms and correctly reason about their $3$D geometry independent of orientation and position.
A powerful method to restrict a neural network to the functions of interest, such as a molecular property, is to exploit the \textit{symmetry} of the data by constraining \textit{equivariance} with respect to transformations from a certain symmetry group \cite{battaglia2018relational, bronstein2021geometric}.

Graph Neural Networks (GNNs) have been applied on a widespread of molecular structures, such as in the prediction of quantum chemistry properties of small molecules \cite{schuett2017schnet, gilmer2017neural} but also on macromolecular structures like proteins \cite{fout, ingraham} due to the natural representation of structures as graphs, with atoms as nodes and edges drawn based on bonding or spatial proximity. These networks generally encode the $3$D geometry in terms of rotationally invariant representations, such as pairwise distances when modelling local interactions which leads to a loss of directional information, while the addition of angular information into network architecture has shown to be beneficial in representing the local geometry \cite{klicpera2020directional}.

Neural models that preserve equivariance when working on point clouds in $3$D space have been proposed \cite{thomas2018tensor, anderson2019cormorant, fuchs2020se3transformers, batzner2021e3equivariant} which can be described as Tensorfield Networks. These physics-inspired models leverage higher-order tensor representations and require additional calculations, to construct the basis for the transformations of their learnable kernels, which can be expensive to compute.\\
While these models enable the interaction between different-order representations, (often referred to as type-$l$ representation), many data types are often restricted to scalar values (type-$0$ e.g., temperature or energy) and $3$D vectors (type-$1$ e.g., velocity or forces). Another choice of using \textit{more} information with the (limited) data at hand and build data-efficient models on point clouds\footnote{An example of more information preservation is when considering relative positions between points in $3$D space, where the information of \textit{orientation} is maintained, as opposed when only the (scalar; invariant) distances between points are considered.}  through equivariant functions is to operate directly on Cartesian coordinates \cite{satorras2021en, schutt2021equivariant, jing2021learning, jing2021equivariant} and \textit{explicitly} define the (equivariant) transformations
which is conceptually simpler and does not require the basis calculations as in Tensorfield Networks.

In this work, we introduce Equivariant Graph Attention Networks (EQGAT) operating on point clouds that can scale up to hundreds of atoms when considering larger systems, such as proteins or protein-ligand complexes but also achieves state-of-art results on the prediction of quantum mechanical properties of small molecules. Our model architecture implements a novel attention-mechanism which is invariant to global rotations and translations of inputs and includes spatial- but also content related information which serves as powerful edge embedding when propagating information in the Message Passing Neural Networks (MPNNs) \cite{gilmer2017neural} framework.

\section{Methods and Related Work}\label{ref:sec-2}
\paragraph{Preliminaries.}
In this work we consider vector spaces for representing a point cloud graph $G=(V, P)$ to a feature space. The point cloud has the vertex set $V = \{v_1, \dots, v_N\}$ and positional set $P=\{\vec{p}_1, \dots\, \vec{p}_N\}$, where $\vec{p}_i \in \R^{3}$ states the Cartesian coordinate of node $i$.
Let $\mathbf{P}\in \R^{N\times 3}$ denote the coordinate matrix of the point-cloud $G$ with an arbitrary ordering $\pi(\cdot)$ along the first axis. A graph adjacency matrix $\mathbf{A}\in \R^{N\times N}$ can be constructed by defining a distance cutoff $c$ and set $a_{ij}=1$ if $d_{ij} = ||\vec{p}_i - \vec{p}_j||_2 < c$ and $0$ else.\\
The (symmetric) Euclidean distance matrix $\mathbf{D}\in \R_{+}^{N\times N}$ can be obtained using the canonical inner product: \begin{equation}\label{eq:euclidean-distance}
    (\mathbf{D})_{i,j} = \sqrt{\vec{p}_i^\top\vec{p}_i + \vec{p}_j^\top\vec{p}_j  - 2\vec{p}_i^\top\vec{p}_j}.
\end{equation}
We aim to develop a Graph Neural Network (GNN) model that transforms feature embeddings invariantly or equivariantly to arbitrary rotations in 3D space. The advantage of including equivariant features into the model as a structural prior lies in the fact that some task-related predicted quantities such as forces $\vec{F}$ in Molecular Dynamics, or the dipole moment $\vec{\mu}$ in the multipole expansion of the electron density \cite{schutt2021equivariant} are \textit{geometric} quantities and transform as order-1 tensor when a rotation is performed on the analyzed rigid body. Even though some properties of interest might in fact not be geometric tensors, i.e., are scalar representations, preserving the information of the geometry within the network architecture, has shown to  be beneficial for their prediction \cite{miller2020relevance, schutt2021equivariant}.  

Formally, in our work we aim to model scalar-valued features $\mathbf{s} \in \R^{F_s}$ and vector-valued features $\vec{\mathbf{v}} \in \R^{3\times F_v}$ in our architecture separately under the constraint that scalar (type-$0$) features transform invariantly under rotations and vector (type-$1$) features transform equivariantly under rotations.

\subsection{Invariance and Equivariance}
Let $X \subset \R^{F_s} \times \R^{3 \times F_v}$ be the considered vector space.
Our feature representation can be expressed as tuple of scalar- and vector components $\mathbf{x} = (\mathbf{s}, \vec{\mathbf{v}}) $, where $\mathbf{s}$ is the invariant and $\vec{\mathbf{v}}$ is the equivariant representation that transforms accordingly when a group action is applied on them. The symmetry considered in this paper is the $\textsc{SO}(3)$ group of orthogonal matrices with determinant $1$ in $\R^{3 \times 3}$, i.e.,
\begin{align}
\textsc{SO}(3)=
\{\mathbf{R} \in \mathbb{R}^{3\times 3}:&  \nonumber \\
\mathbf{R}^\top \mathbf{R} = \mathbf{R}\mathbf{R}^\top = \mathbf{I}_3, ~~\text{det}(\mathbf{R}) = 1\}    
\end{align}
    
Let $T_g: X \xrightarrow{}{} X$ be a set of transformations on $X$ for the group $ g\in \textsc{SO}(3)$. 
The action of $g=\mathbf{R}\in \textsc{SO}(3)$ on the vector space $X$ is defined as:
\begin{equation}\label{eq:so3-action}
    T_g(\mathbf{x}) = (\mathbf{s}, \mathbf{R}\vec{\mathbf{v}})~.
\end{equation}
The group action in Eq. \eqref{eq:so3-action} reveals that scalar component $\mathbf{s}$ is not affected by the rotation, as this is not a geometric quantity in $\R^3$ and therefore invariant\footnote{To be precise, the group action applied on the scalar embedding is the \textit{trivial} representation, which is represented as the identity map through the $F_s$-dimensional unit matrix.}. The vector feature $\vec{\mathbf{v}}$ however, transforms as an order-$1$ tensor and is therefore rotated.

We say a function $f: X \xrightarrow{}{} Y$ is equivariant to $g$ if there exists and equivalent transformation on its output space $S_g: Y \xrightarrow[]{} Y$ such that following holds:
\begin{equation}\label{eq:equivariant_functions}
    f(T_g(\mathbf{x})) = S_g(f(\mathbf{x}))~.
\end{equation}
We note that a composition of equivariant functions satisfying Eq. \eqref{eq:equivariant_functions} is equivariant again. In our work, $f$ is a message passing (MP) layer \cite{gilmer2017neural} that updates node feature representation $\{\mathbf{x}_i\}_{i=1}^N$ of the point cloud in an iterative manner such that
\begin{equation}\label{eq:message-passing-equiv}
    \mathbf{x}^{(l+1)} = (\mathbf{s}^{(l+1)}, \vec{\mathbf{v}}^{(l+1)}) = f((\mathbf{s}^{(l)}, \vec{\mathbf{v}}^{(l)})) = f(\mathbf{x}^{(l)})~.
\end{equation}
Combining the group action in Eq. \eqref{eq:so3-action} with the equivariance property of the function $f$ in \eqref{eq:equivariant_functions} results into
\begin{align}\label{eq:group-action-mp}
    f(T_g(\mathbf{x}^{(l)})) &= S_g(f(\mathbf{x}^{(l)})) \nonumber \\
    &= S_g(\mathbf{x}^{(l+1)})= (\mathbf{s}^{(l+1)},\mathbf{R}\vec{\mathbf{v}}^{(l+1)})~,
\end{align}
which shows that the group action $g$ commutes with the function $f$.

\paragraph{Tensor / Outer Product.}
The tensor product\footnote{Often also referred to as Outer or Kronecker product.} $\otimes$ between two vectors $\mathbf{a} \in \R^n$ and $\mathbf{b} \in \R^{k}$ is computed as 
\begin{equation}\label{eq: tensor-product}
    \mathbf{a} \otimes \mathbf{b} = (\mathbf{a} \mathbf{b}^\top) \in \R^{n \times k}~, 
\end{equation}
and returns a matrix given two vectors. The tensor/outer product will be a useful operation to construct equivariant features by combining type-$1$ and type-$0$ representations.

\subsection{Message Passing Neural Networks (MPNNs)}
MPNNs \cite{gilmer2017neural} construct complex representations of vertices within their local neighbourhood through an iterative exchange of messages followed by updates of vertex features. Since MPNNs utilize shared trainable layers among nodes, permutation equivariance is preserved. As mentioned in Section \ref{ref:sec-2}, edges between vertices of the point cloud are specified by their relative position $\vec{p}_{ij} = \vec{p}_j - \vec{p}_i$ of vertices $i,j$ within a local neighbourhood through a distance cutoff $c>0$ , i.e., $\mathcal{N}(i)=\{j: ||\vec{p}_{ij}||_2 = d_{ij} < c\}$. 
The (Euclidean) distance function in \eqref{eq:euclidean-distance} is an invariant function 
as the canonical inner product between two vectors $\vec{p}_i$ and $\vec{p}_j$ after the group action $\mathbf{R}\in \textsc{SO}(3)$ does not change, i.e., $(\mathbf{R}\vec{p}_i)^\top (\mathbf{R}\vec{p}_j) = (\vec{p}_j^\top \mathbf{R}^{\top})(\mathbf{R}\vec{p}_j) = \vec{p}_i^\top \mathbf{I}_3 \vec{p}_j = \vec{p_i}^\top \vec{p}_j$. Hence, the inner product, or a composition of it, like the norm, is a natural way to obtain $\textsc{SO}(3)$ invariant features from equivariant features. The usage of the \textit{tensor}/\textit{outer} product and \textit{inner} product to obtain equivariant and invariant features, respectively, will be explained in the next Section.\\

Standard MPNNs implement a \textit{message} and \textit{update} function for feature representation as
\begin{align}
    \label{eq:message-function}
    &\mathbf{m}_{i}^{(l+1)} = \sum_{j\in \mathcal{N}(i)} f_m^{(l)}(\mathbf{x}_i^{(l)} \mathbf{x}_j^{(l)}, \vec{p}_{ji}) = \sum_{j\in \mathcal{N}(i)} \mathbf{m}_{ji}^{(l)}~, \\
    \label{eq:update-function}
    &\mathbf{x}_{i}^{(l+1)} = f_u^{(l)}(\mathbf{x}_i^{(l)}, \mathbf{m}_i^{(l+1)})~.
\end{align}
In our paper, we aim to implement the message ($f_m$) and update ($f_u$) function to be equivariant, i.e., to satisfy the property in Equation \eqref{eq:group-action-mp}.

\begin{figure*}[ht!]
    \centering
    \includegraphics[width=\textwidth]{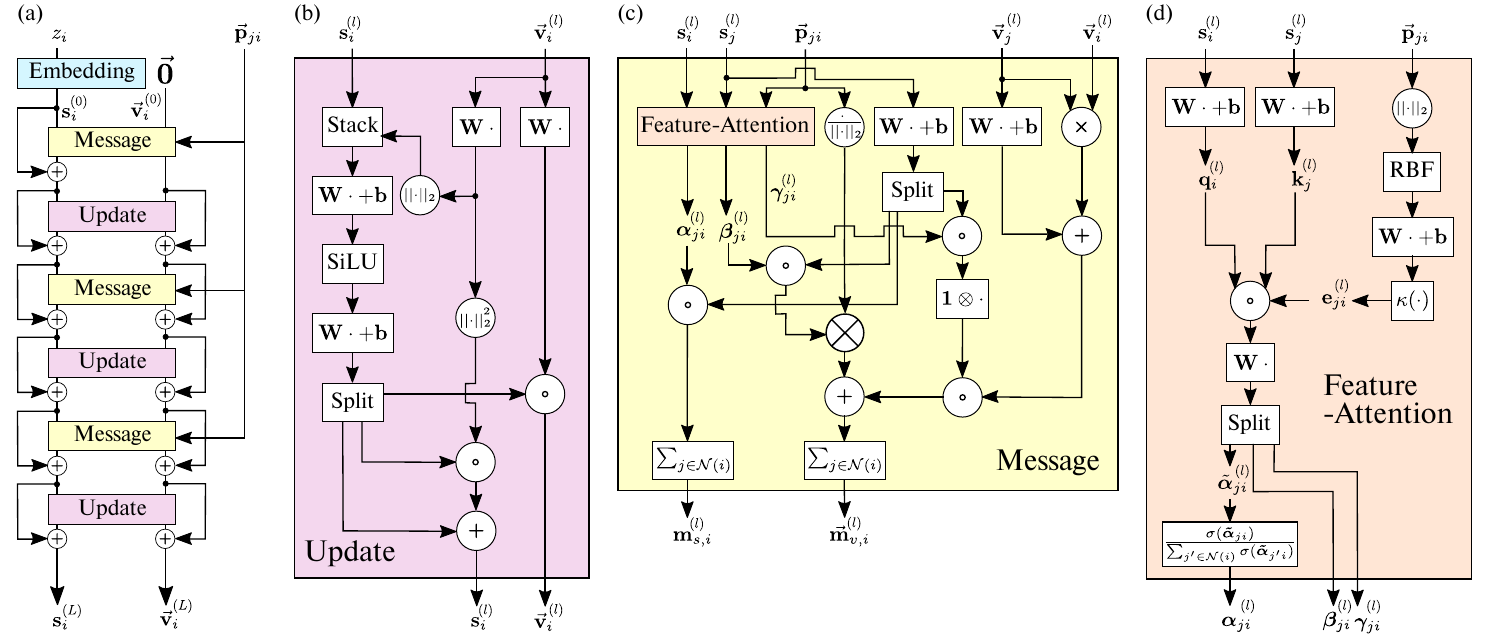}
    \caption{ \textbf{(a)} Overview of the EQGAT encoder architecture with initial scalar features obtained through an embedding layer. \textbf{(b)} Update layer to enable information flow between scalar- and vector features. \textbf{(c)} Message function that creates vector features through the tensor product of relative positions with filtered scalar features combined with existing filtered vector features. Scalar messages are obtained via filtered value tensors. \textbf{(d)} Feature Attention calculation that inputs SO(3) invariant representations and outputs a content as well as spatial based filter.}
    \label{fig:model-architecture}
\end{figure*}

\subsection{Related Work}
Neural networks that specifically achieve E($3$) or SE($3$) equivariance have been proposed in Tensorfield Networks (TFN) \cite{thomas2018tensor} and its variants in the covariant Cormorant \cite{anderson2019cormorant}, NequIP \cite{batzner2021e3equivariant} and SE(3)-Transformer \cite{fuchs2020se3transformers}. With TFNs, equivariance is achieved through the usage of equivariant function spaces such as Spherical Harmonics combined with Clebsch-Gordan coefficients, while others resort to lifting the spatial space to higher-dimensional spaces such as Lie group spaces \cite{ finzi2020generalizing}. Since no restriction on the order of tensors is imposed on these methods, sufficient expressive power of these models is guaranteed, but at a cost of excessive computational calculations with increased time and memory.
To circumvent the expensive computational cost, another line of research
proposed to directly implement equivariant operations in the original (Cartesian) space, providing and efficient approach to preserve equivariance without performing complex space transformations as introduced in the E($n$)-GNN \cite{satorras2021en}, GVP \cite{jing2021learning, jing2021equivariant} and PaiNN \cite{schutt2021equivariant} architectures.

Our proposed model implements equivariant operations in the original Cartesian space and includes a continuous filter through the self-attention coefficients which serve as spatial- and content based edge embedding in the message propagation, as opposed to the PaiNN model where the filter solely depends on the distance. The E($n$)-GNN architecture does not learn higher-dimensional type-$1$ vector features, but only updates given type-$1$ features through a weighted linear combination of such, where the (learnable) scalar weights are obtained from invariant embeddings.  
The GVP model which was initially designed to work on macromolecular structures includes a complex message functions of concatenated node- and edge features composed with a series of GVP-blocks that enables information exchange between type-$0$ and type-$1$ features, with a potential disadvantage of discontinuities through non-smooth components for distances close to the cutoff.\\
A concurrent work by \cite{thoelke2022equivariant} proposed the Equivariant-Transformer (ET),  which is similar to ours, and obtains strong results on QM9 \cite{qm9}, MD17 \cite{md17} and ANI-1 \cite{ani} datasets. Within their model, to initialize node embeddings, they implement an additional neighbouring embedding $n_i$ for target node $i$, which can thought of a continuous filter convolution as proposed in the SchNet architecture \cite{schuett2017schnet}. This neighbouring embedding $n_i$ is then combined with a self-embedding, generally resembling a graph convolutional layer, which our model does not incorporate. Our model differs in a distinguished message function where we implement a feature-attention embedding that aims to filter scalar-\textit{channels} from neighbouring nodes. Furthermore, our message function incorporates an intermediate embedding that is content- as well as spatial dependent to modulate the vector-\textit{channels}, while their filter solely depends on spatial information via interatomic distances. As we benchmark our model on large molecular structures, we assign different channel sizes, $F_s, F_v$ for scalar and vector features, respectively, to be able to train deep models on learning problems that include proteins.

\section{Equivariant Graph Attention Networks}
\label{ref:EQGTNN-start}
The traditional Transformer model as introduced by Vaswani et al. \cite{vaswani2017attention} has revolutionized the field of Natural Language Processing (NLP) \cite{devlin2019bert, dai2019transformerxl} and image analysis \cite{ramachandran2019standalone, zhao2020exploring, dosovitskiy2021image}. The \textit{self-attention} module which lies in the core of Transformers, initially designed to operate on sequence of tokens in NLP, consists in essence of two components: input-dependent \textit{attention-weights} between any two elements of the set, and an embedding for each set-element, called \textit{value}. Since graphs, are naturally represented as sets with underlying structure through the (sparse) connectivity between vertices, the implementation of self-attention, following the MPNN paradigm, was introduced in Graph Attention Networks (GATs) \cite{velickovic2018graph}. Transformer-like GNNs (that work on a fully-connected 2D graph) were recently introduced in \cite{kreuzer2021rethinking, ying2021transformers} and have found success in several graph-learning benchmarks due to their incorporation of (learnable) relative positional encodings into the attention function. \\
Our proposed Equivariant Graph Attention Networks (EQGAT) operates in $3$D space and implements the message passing for each target node $i$ on its local neighbourhood $\mathcal{N}(i)$ as defined in Section \ref{ref:sec-2} to avoid the quadratic complexity of the vanilla self-attention when one target node would interact with all other nodes in the point cloud.  We emphasize that the integration of local neighbourhoods manifests as a powerful inductive bias and in a bio-chemistry context, coincides with the assumption that a large part of energy variations can be attributed to local interactions, although the influence and importance of non-local effects in machine-learned force-fields has been recently analyzed in \cite{spookynet}.\\
In the following Subsections, we will introduce the function components of our EQGAT network displayed in Figure \ref{fig:model-architecture}a and highlight its invariance and equivariance properties.
\subsection{Input Embedding}\label{ref:init-embedding}
We initially embed atoms of small molecules or proteins based on their nuclear charge $Z$. As nuclear charges are discrete and bounded, we utilize a trainable embedding look-up table commonly used in NLP to map the $i-$th atom charge $z_i$ to a feature representation $\mathbf{s}_{i}^{(0)} \in \mathbb{R}^{F_s}$ as
\begin{equation}\label{eq:init-embedding}
    \mathbf{s}_{i}^{(0)} = \text{embed}(z_i),
\end{equation}
which provides a starting (invariant) scalar representation of the node prior to the message passing.\\
As in most cases, no directional information for atoms are available, we initialize the vector features as zero tensor $\vec{\mathbf{v}}_i^{(0)} = \vec{\mathbf{0}} \in \mathbb{R}^{3 \times F_v}$.
\subsection{Distance Encoding}
Interatomic distances $d_{ji} = ||\Vec{p}_{ji}||_2 \in \mathbb{R}_{+}$ are embedded using the Bessel radial basis function (RBF) as introduced by \citeauthor{klicpera2020directional} into a representation $\mathbf{e}^{\text{RBF}}(d_{ji}) \in \mathbb{R}^K$ that serves as distance encoding into the attention mechanism. The Bessel RBF function reads
\begin{align*}
    {e}^{\text{RBF}}_{k}(d_{ji}) = \sqrt{\frac{2}{c}} \frac{\sin{(\frac{k\pi}{c}}d_{ji})}{d_{ji}},
\end{align*}
where $c$ is the distance cutoff and $k=1,\dots, K$. In similar fashion to the continuous filter convolution in the SchNet architecture \cite{schuett2017schnet}, the deterministic radial basis encoding is further transformed using a trainable linear layer to obtain an edge embedding $\tilde{\mathbf{e}}_{ji}:= \tilde{\mathbf{e}}(d_{ji})\in \mathbb{R}^{F_s}$,
\begin{align*}
    \tilde{\mathbf{e}}_{ji} = \mathbf{W}_{e} \mathbf{e}^{\text{RBF}}(d_{ji}) + \mathbf{b}_e~.
\end{align*}
As the final edge embedding $\mathbf{e}_{ji}$ ought to transition smoothly to $\mathbf{0}$ to avoid discontinuities when $d_{ji} \xrightarrow[]{} c$ , we apply a cosine-cutoff function $\kappa: \mathbb{R}_{+}\mapsto [0, 1]$ as introduced in \cite{behler}
\begin{align*}
    \kappa(d_{ji}) = \frac{1}{2}\left(\cos{(\frac{\pi d_{ji}}{c})} + 1\right) \cdot \mathbbm{1}[d_{ji} \leq c]~,
\end{align*}
which leads to the final edge embedding:
\begin{align}\label{eq:final-edge-embedding}
    \mathbf{e}_{ji} &:= \mathbf{e}(d_{ji})\\ \nonumber
    &= \kappa(d_{ji}) \cdot \tilde{\mathbf{e}}_{ji} = \kappa(d_{ji}) \cdot (\mathbf{W}_{e} \mathbf{e}^{\text{RBF}}(d_{ji}) + \mathbf{b}_e)~.
\end{align}
The edge-embedding $\mathbf{e}_{ji} \in \mathbb{R}^{F_s}$ in Equation \eqref{eq:final-edge-embedding} can be interpreted as a convolutional filter that solely depends on an invariant representation, i.e., the distance between the nodes $j$ and $i$.

\subsection{Feature Attention}\label{sec:feature-attention}
Our attention mechanism depends on invariant scalar representations and includes the distance encoding in Eq. \eqref{eq:final-edge-embedding} to incorporate spatial connectivity. A novelty of our approach is that the attention coefficient between two vertices $j$ and $i$ is in fact obtained per feature-channel instead for the entire embedding. Formally, given the invariant scalar embeddings $\mathbf{s}_i, \mathbf{s}_j \in \mathbb{R}^{F_s}$ we use two linear layers to obtain query- and key embeddings as
\begin{align}\label{eq:query-key}
    \mathbf{q}_i &= \mathbf{W}_q \mathbf{s}_i + \mathbf{b}_q~ \in \mathbb{R}^{F_s}~, \nonumber\\
    \mathbf{k}_j &= \mathbf{W}_k \mathbf{s}_j + \mathbf{b}_k~ \in \mathbb{R}^{F_s}~.
\end{align}
We proceed to compute a content \textit{and} spatial dependent anisotropic embedding through the elementwise product as
\begin{equation}
    \tilde{\mathbf{a}}_{ji} = \mathbf{q}_i\odot \mathbf{k}_j \odot \mathbf{e}_{ji}~.
\end{equation}
The embedding vector $\tilde{\mathbf{a}}_{ji} \in \mathbb{R}^{F_s}$ contains both, semantic information through the query and key representations as well as spatial information via the relative positional encoding expressed through the edge embedding.\\
Next, we utilize another weight matrix $\mathbf{W}_{a}\in \mathbb{R}^{(F_s + 2F_v) \times F_s}$ to obtain an embedding vector to compute the feature-attention and filter components for the equivariant features,
\begin{equation}\label{eq:feature_attention_all}
    \mathbf{a}_{ji} = \mathbf{W}_{a} \tilde{\mathbf{a}}_{ji} = [\tilde{\bm{\alpha}}_{ji}, \bm{\beta}_{ji}, \bm{\gamma}_{ji}] \in \mathbb{R}^{F_s + 2F_v}~.
\end{equation}
The feature-attention is calculated using the $\bm{\tilde{\alpha}}_{ji}$ tensor as follows
\begin{equation}\label{eq:feature_attention_scalar}
    \bm{\alpha}_{ji} = \frac{\sigma(\bm{\tilde{\alpha}}_{ji})}{\sum_{j' \in \mathcal{N}(i)} \sigma(\bm{\tilde{\alpha}}_{j'i})}~\in (0,1)^{F_s},
\end{equation}
where $\sigma(\cdot)$ is the sigmoidal activation function and applied componentwise for each channel.

\subsection{Value Transforms}
Another constituent of the self-attention mechanism is the pointwise transform of neighbouring (source) nodes which will be multiplied with the output of the feature-attention mechanism described in Eq. \eqref{eq:feature_attention_all}. Value tensors for scalar- and vector features are computed using weight matrices $\mathbf{W}_{sv} \in \mathbb{R}^{F_s \times (F_s + 2F_v)}$, and $\mathbf{W}_{vv} \in \mathbb{R}^{F_v \times F_v}$.
\begin{align}\label{eq:value-transforms}
    &\mathbf{v}_{s, j} = \mathbf{W}_{sv} \mathbf{s}_j + \mathbf{b}_{sv}~, \nonumber \\
    &\vec{\mathbf{v}}_{v, j} = \vec{\mathbf{v}}_j \mathbf{W}_{vv}, 
\end{align} where the product on $\vec{\mathbf{v}}_j \in \R^{3 \times F_v}$ is applied on the feature axis. By such application of the linear transformation in Eq. \eqref{eq:value-transforms}, the equivariance property in Eq. \eqref{eq:so3-action} is maintained.\\
The rationale behind the additional $2F_v$ output channels for the scalar embedding $\mathbf{v}_{s,j}$ is associated to the \textit{construction} of equivariant features through the tensor product with invariant features as will be explained later.\\
We split the value tensor of the scalar part $\mathbf{v}_{s, j}$ into three tensors $[\mathbf{v}_{s_0,j}, \mathbf{y}_{v_0, j}, \mathbf{y}_{v_1, j}]$.

\paragraph{Self-Attention as Continuous Filter Convolution.}
The invariant feature-attention coefficients in Eq. \eqref{eq:feature_attention_scalar} depend on content related as well as spatial based information and act as a filter when elementwise multiplied with the pointwise transforms of scalar-value tensors $\mathbf{v}_{s_0, j}$ to obtain the scalar message-embedding $\forall j \in \mathcal{N}(i)$:
\begin{equation}\label{eq:scalar_message_embedding-0}
    \mathbf{m}_{s, ji} = \bm{\alpha}_{ji} \odot \mathbf{v}_{s_0, j}~.
\end{equation}
Furthermore, we compute two additional elementwise products to obtain filtered message tensors:
\begin{align}\label{eq:scalar_message_embedding-1}
    \mathbf{m}_{v_0, ji} = \bm{\beta}_{ji} \odot \mathbf{y}_{v_0, j}~, \nonumber \\
    \mathbf{m}_{v_1, ji} = \bm{\gamma}_{ji} \odot \mathbf{y}_{v_1, j}~. 
\end{align}
We highlight that the message embeddings in \cref{eq:scalar_message_embedding-0,eq:scalar_message_embedding-1} are calculated using $\textsc{SO}(3)$ invariant representations through the initial query-key computation in \eqref{eq:query-key} and only the channels of the $\alpha$-coefficients are bounded in the unit interval to obtain "modulated" scalar representations when multiplied with the $\mathbf{v}_{s_0,j}$ tensor.

Our model differs from the SE(3)-Transformer proposed by \citeauthor{fuchs2020se3transformers}
as we do not rely on Spherical Harmonics calculation and Clebsch-Gordan decomposition
to build equivariant functions, but we \textit{explicitly} design functions that are equivariant and operate on 3D-Cartesian coordinates for faster and more memory-efficient calculations as described in the next paragraph. Additionally, our proposed model aims to decouple the information flow between scalar- and vector representations within the attention-mechanism by just using scalar/invariant representations $\mathbf{a}_{ji}$ in Equation \eqref{eq:feature_attention_all} to modulate scalar and vector embeddings, while the (later applied) update function is used to enable an information flow between vector representations to scalar embeddings.\\
An important fact is that the attention coefficient(s) which serve as a filter are required to be SO($3$) invariant. Such requirement originates from the idea that the attention coefficient is multiplied with a type-1 feature\footnote{When considering the vector features.} which itself transforms, when a group action (in our case a Rotation), is applied.
For the case that one would want to include type-1 features into the attention calculation, as proposed by \cite{fuchs2020se3transformers} through additional type-1 related query and key representations, a contraction along the spatial axis, such as an inner product between those pairs is required to obtain SO(3) invariance. We tried such construction in our initial development of the attention function, but found worse performance in validation and additional computational complexity.
\paragraph{Building Equivariant Features.}
In case that no initial vector features such as velocity, or forces are available, equivariant representations in the point cloud can be constructed as a function of (relative) position $\vec{p}_{ij} = \vec{p}_j - \vec{p}_i$. Relative position vectors in 3D space can have unbounded norms, so a common practice is to obtain relative positions with unit norm $\vec{p}_{e, ij} = \frac{\vec{p}_{ij}}{||\vec{p}_{ij}||_2}$ 
which describe points on the $2-$dimensional unit sphere $S^{2} = \{\vec{r} \in \R^3: ||\vec{r}||_2 = 1\}$.\\
Equivariant interactions between node $j$ and $i$ are modelled through the tensor-product (\textit{cf}. Eq. \eqref{eq: tensor-product}) of the equivariant representation $\vec{p}_{e, ji} \in \R^{3}$ with invariant representation $\mathbf{m}_{v_{0}, ji} \in \R^{F_v}$.\\
Formally, equivariant features are obtained as 
\begin{equation}\label{eq:equivariant_feature_creation-0}
    \vec{\mathbf{y}}_{0, ji} = \vec{p}_{e, ji} \otimes \mathbf{m}_{v_{0}, ji} = \vec{p}_{e, ji} \mathbf{m}_{v_{0}, ji}^\top    ~~\in \R^{3 \times F_v},
\end{equation}
and (hidden) equivariant type-$1$ embeddings are filtered using the (elementwise) scalar multiplication
\begin{equation}\label{eq:equivariant_feature_creation-1}
    \vec{\mathbf{y}}_{1, ji} = (\mathbf{1} \otimes \mathbf{m}_{v_1, ji}) \odot  \vec{\mathbf{a}}_{v, ji} = (\mathbf{1} \mathbf{m}_{v_1, ji}^\top) \odot  \vec{\mathbf{a}}_{v, ji}~,
\end{equation}
where $\mathbf{1} \in \R^3$ with $1$'s in the components and $\vec{\mathbf{a}}_{v, ji}$ is obtained through the cross product $\times$ as:
\begin{align*}
    \vec{\mathbf{a}}_{v, ji} = \vec{\mathbf{v}}_{j} \times \vec{\mathbf{v}}_{i} + \vec{\mathbf{v}}_{v, j}~.
\end{align*}
The rationale for including the cross product of the two transformed vector features is to enable interaction between type-1 features and is inspired from TFNs \cite{thomas2018tensor} which models the tensor product between two type-1 vectors resulting in a rank-2 tensor. Such rank-2 tensor includes antisymmetric elements presented in the cross product.
We combine the two equivariant representations into a final aggregated equivariant message embedding
\begin{align}\label{eq:aggr_message_vector}
    \vec{\mathbf{m}}_{v, i} = \sum_{j \in \mathcal{N}(i)} (\vec{\mathbf{y}}_{0, ji} + \vec{\mathbf{y}}_{1, ji})~.
\end{align}
The aggregated invariant message embedding is obtained in the same fashion, using filtered messages (see. Eq. \eqref{eq:scalar_message_embedding-0}) between target node $i$ and its neighbours
\begin{equation}
    \mathbf{m}_{s, i} = \sum_{j \in \mathcal{N}(i)}\mathbf{m}_{s, ji}~.
\end{equation}
We follow our vector space terminology and write
\begin{align*}\label{eq:final-aggregated-message}
    \mathbf{m}_{i} = (\mathbf{m}_{s,i}, \vec{\mathbf{m}}_{v, i}) \in \R^{F_s} \times \R^{3 \times F_v}~,
\end{align*}
as the final aggregated message embedding (\textit{cf}. Eq. \eqref{eq:message-function}) which by design also satisfies the equivariance property in Eq. \eqref{eq:group-action-mp}. We prove this claim in the Appendix \ref{sec:appendix-rotation-equiv}.\\
At this point, it is worth mentioning that the tensor product applied in \cref{eq:equivariant_feature_creation-0,eq:equivariant_feature_creation-1} reduces to a scalar multiplication in case the (invariant) embedding(s) $\mathbf{m}_{v, ji}$ are one-dimensional as in the E($n$)-GNN architecture \cite{satorras2021en}. Furthermore, notice that by such construction, higher-dimensional type-$1$ embeddings can also be constructed in $\R^n$ instead of $\R^3$.
\subsection{Update}
We update the hidden embedding for each target node $i$ by adding the aggregated message $\mathbf{m}_i$ with the previous hidden state
\begin{equation}\label{eq:intermediate-update}
    \mathbf{\tilde{x}}_i \xleftarrow{} \mathbf{x}_i + \mathbf{m}_i~.
\end{equation}
The state $\mathbf{\tilde{x}}_i$ in Equation \eqref{eq:intermediate-update} includes complex transformation performed in the $\mathbf{m}_i$ embedding via the attention mechanism, while \textit{no} self-interaction transformation on $\mathbf{x}_i \in \R^{F_s}\times \R^{3 \times F_v}$ has been applied. We propose to first update the $i-$th's hidden state in a residual-connection manner and then apply a pointwise \textit{update}-layer similar as in the PaiNN architecture \cite{schutt2021equivariant} with the use of gated equivariant non-linearities \cite{weiler2018learning} to combine invariant and equivariant information in the $\tilde{\mathbf{x}}$ representation to update its state (\textit{cf}. Eq. \eqref{eq:update-function}) into $\mathbf{x}_i \xleftarrow[]{} f_u(\tilde{\mathbf{x}}_i)$ as illustrated in Figure \ref{fig:model-architecture}b.

\section{Experiments and Results}

\begin{table}[ht!]
\caption{QM$9$ Mean Absolute Error (MAE) on the test set.\\
Models displayed with \text{*} use different (random) train/val and test splits. For our EQGAT, we report averaged MAEs over 3 runs.}
\begin{center}
\resizebox{\columnwidth}{!}{%
    \begin{tabular}{lcccccc}
        \hline
         Tasks & $\alpha$ & $\Delta \epsilon$ & $\epsilon_{\textsc{HOMO}}$& $\epsilon_{\textsc{LUMO}}$ & $\mu$ & $C_{\nu}$ \\
         Units & bohr$^3$ & meV & meV & meV & $D$ & $\frac{\text{cal}}{\text{mol K}}$ \\
        \hline
        NMP & .092 & 69 & 43 & 38 & .030 &  .040  \\
        SchNet$^*$ & .235  & 63 & 41 & 34 & .033 &  .033  \\
        Cormorant & .085 & 61 & 34 & 38 & .038 &  .026  \\
        LieConv & .084 & 49 & 30 & 25 & .032 &  .038 \\
        DimeNet++$^{\text{*}}$ & .044 & 33 & 25 & 20 & .030 &  .023  \\
        SE(3) Tr. & .142 & 53 & 35 & 33 & .051 & .054    \\
        E($n$)-GNN & .071 & 48 & 29 & 25 & .029 &  .031 \\
        ET$^*$& .010 & 38 & 21 & 18 & .002 &  .026 \\
        PaiNN$^{\text{*}}$ & .045 & 46 & 28 & 20 & .012 & .024 \\
        \hline
        EQGAT & .063 & 44 & 26 & 22 & .014 & .027 \\
        EQGAT$^{*}$& .053 & 32 & 20 & 16 & .011 & .024 \\
        \hline
    \end{tabular}
    }%
\end{center}
\label{tab:QM9-results}
\end{table}

We test the effectiveness of our proposed EQGAT model on four publicly available molecular benchmark datasets which pose significant challenges for the development of efficient and accurate prediction models in small molecules drug discovery but also protein design on different data scales.
\subsection{QM9}
The QM9 dataset \cite{qm9} is a chemical property regression benchmark and consists of $134$k small molecules with up to $29$ atoms per molecule. Molecules are represented as point clouds with each atom having a $3$D position and a five dimensional one-hot encoding that describes the atom type (H, C, N, O, F), and additional features that can be derived from the 2D topological graph, such as bond-types. In our experiments, we only use the atom positions and atom types/charges as input features for our EQGAT model as described in the Methods Section \ref{ref:init-embedding}. To compare our method with the literature, we import the training, validation- and test-splits from \cite{anderson2019cormorant}
which consists of $100$k, $18$k and $13$k compounds, respectively.
We adopt the hyperparameters from \cite{satorras2021en} and implement a $7$-layer EQGAT-encoder with $F_s=128$ scalar and $F_v=32$ vector channels as well as $K=20$ radial basis functions to encode interatomic distances. For the downstream networks we only utilize scalar embeddings from the last hidden layer.
For a detailed architecture of the entire network with approximately $1.1$M trainable parameters, we refer to Section \ref{sec:appendix-qm9} of the Appendix. \\
We optimized and report the Mean Absolute Error (MAE) between prediction and ground truths on 6 targets and compare to NMP \cite{gilmer2017neural}, SchNet  \cite{schuett2017schnet}, Cormorant  \cite{anderson2019cormorant}, LieConv \cite{finzi2020generalizing}, DimeNet++ \cite{klicpera2020directional}, 
SE(3)-Transformer \cite{fuchs2020se3transformers}, E$(n)$-GNN \cite{satorras2021en}, ET \cite{thoelke2022equivariant} and PaiNN \cite{schutt2021equivariant}.\\

\begin{table}[ht!]
\caption{Benchmark results on three ATOM3D tasks.\\
We report the results for the Baseline models from \cite{townshend2021atom3d}.
For the E($n$)-GNN, PaiNN and our EQGAT model, we report averaged metrics over 3 runs.
}
\begin{center}
\resizebox{\columnwidth}{!}{%
    \begin{tabular}{l|cc|cc|c}
        \hline
         Tasks & \multicolumn{2}{c|}{PSR $(\uparrow)$} & \multicolumn{2}{c|}{RSR $(\uparrow)$} & LBA $(\downarrow)$\\
         Metric & Global $R_S$ & Mean $R_S$ & Global $R_S$ & Mean $R_S$ & RMSE  \\
         \hline
        CNN & 0.431 & 0.789 & {0.264}  & 0.372 & \textbf{1.415} \\
        GNN & {0.515} & 0.755 & 0.234  & \textbf{0.512} & 1.570 \\
        GVP-GNN & 0.511 & {0.845} & 0.211 & 0.330 & 1.594 \\
        E($n$)-GNN & 0.466 & 0.789 & - & - & 1.558 \\
        PaiNN & 0.485 & 0.808 & - & - & 1.548 \\
        \hline
        EQGAT & \textbf{0.576} & \textbf{0.849} & \textbf{0.322} & 0.365 & 1.489 \\
        \hline
  \end{tabular}
    }%
\end{center}
\label{tab:Atom3D-results}
\end{table}

\textbf{Results} of our proposed model are reported in Table \ref{tab:QM9-results}. Our EQGAT architecture obtains very strong performance on the 6 predicted targets compared to recent state-of-the-art models while our model similar to the E($n$)-GNN, ET, and PaiNN model implements equivariant functions in the original Cartesian space without the need of higher-order tensor representations. We believe that the implementation of \textit{output-specific} layers for the prediction of certain targets such as the magnitude of the dipole moment $\mu$ or the electronic spatial extent $\langle R^2 \rangle$ (not listed in Table \ref{tab:QM9-results}) as proposed by \cite{schutt2021equivariant} using the GNN encoder's scalar and vector features $\mathbf{x}^{(L)}=(\mathbf{s}^{(L)}, \Vec{\mathbf{v}}^{(L)})$ in combination with the global geometry via Cartesian coordinates $\{\Vec{p}_i\}_{i=1}^N$ might lead to better performance on those targets. As we adopt the overall architecture for all 6 targets, we did not implement output-specific (decoder) layers for each target. In comparison to the concurrent ET architecture \cite{thoelke2022equivariant}, our model obtains similar results on the 6 targets while being parameter-lighter ($1.1$M vs. $6.9$M).

We refer the reader to the Appendix \ref{sec:appendix-qm9} for the architectural details and performance of our models on all $12$ QM$9$ targets including energy predictions.

\subsection{ATOM3D}
The ATOM3D benchmark \cite{townshend2021atom3d} is a collection of eight tasks and datasets for learning on atomic-level 3D molecular structures of different kinds, i.e., proteins, RNAs, small molecules and complexes. Since proteins perform specific biological functions essential for all living organisms and
hence, play a key role when investigating the most fundamental questions in the life sciences, we focus our experiments on the learning problems often encountered in structural biology with different difficulties due to data scarcity and varying structural sizes.
We use provided training, validation and test splits from ATOM3D and refer the interested reader to the original work \cite{townshend2021atom3d} for more details.
For all benchmarks, we compare against the Baseline CNN and GNN models provided by the authors from ATOM3D, the GVP-GNN reported in \cite{jing2021equivariant} and PaiNN \cite{schutt2021equivariant} as well as the E($n$)-GNN \cite{satorras2021en} architectures using our own implementations.\\
All of our implemented models utilize a 5-layer GNN encoder with $128$ (scalar) channels as hidden invariant embedding. For our EQGAT model we apply $16$ vector channels, while the PaiNN architecture models $128$ vector channels and the E($n$)-GNN does not include learnable vector-feature but just updates the positional coordinates. We refer the reader to the Appendix \ref{sec:appendix-atom3d} for a detailed description concerning the implementation of the networks on the tasks.  

The Protein and RNA Structure Ranking tasks (PSR / RSR) in ATOM3D are both regression tasks with the objective to predict the quality score in terms of \textit{Global Distance Test} (GDT\_TS) or Root-Mean-Square Deviation (RMSD) for generated Protein and RNA models wrt. to its experimentally determined ground-truth structure.
Being able to reliably rank a biopolymer structure requires a model to accurately learn the atomic environments such that discrepancies between a ground truths state an its corrupted version can be distinguished. We evaluated our model on the biopolymer ranking and obtained the best results on the current benchmark, reported in Table \ref{tab:Atom3D-results} in terms of Spearman rank correlation. Our proposed model performs particularly well on the PSR task outperforming the GVP-GNN \cite{jing2021equivariant}. We noticed that the RSR benchmark was particularly difficult to validate as only a few dozen experimentally determined RNA structures are existent to date, and the structural models generated in the ATOM3D framework are labeled with the RMSD to its native structure, which is known to be sensitive to outlier regions, for exampling by inadequate modelling of loop regions \cite{gl_ds}, while the GDT\_TS metric might be a better suited target to predict a ranking for generated RNA structures as in the PSR benchmark. In our experiments, we tried to obtain comparable results for the PaiNN and E($n$)-GNN architecture on the biopolymer benchmarks but found it difficult to make our implementation train on these two benchmarks after careful comparison to the authors original source code.

Another challenging and important task for drug discovery projects is estimating the binding strength (affinity) of a candidate drug atomistic's interaction with a target protein. We use the ligand binding affinity (LBA) dataset and found that among the GNN architectures, our proposed model obtains the best results, while also being computationally cheap and fast to train. The best performing model in the LBA-task is a 3D CNN model which works on the joint protein-ligand representation using voxel space and enforcing equivariance through data augmentation. The inferior performance of all equivariant GNNs might be caused by the need of larger filters to better capture the locality, where 3D CNNs have an advantage when using voxel representations. Furthermore, as all GNN models jointly represent ligand- and protein as \textit{one} graph by connecting vertices through a distance cutoff of $4.5\angstrom$, we believe that such union leads to an information loss of distinguishing the atom identity from the ligand and protein. A promising direction to investigate is to incorporate a ligand and protein GNN encoder seperately and merge the two embeddings prior the binding affinity prediction, similar to Graph Matching Networks \cite{li2019graph}.\\

\textbf{Computational Efficiency.}
We assess the computational efficiency of the proposed equivariant Graph Attention Network and compare it against our implemented PaiNN and E($n$)-GNN architectures on the PSR, RSR and LBA benchmarks from ATOM3D. As these datasets consist of graphs with up to hundreds of atoms, computationally- and memory efficient models are preferred such that batches of graphs can be stored on GPU memory and trained fast.
We measure the inference time of a random batch comprising 10 macromolecular structures on an NVIDIA V$100$ GPU and observe the fastest execution time for our proposed model on the LBA dataset with $23.67$ms against the PaiNN and E($n$)-GNN with $30.75$ms and $25.76$ms, respectively.\newline
Most notably on the PSR and RSR datasets which consist of large biopolymer structures, the time difference is more obvious for EQGAT with $42.79$ms against PaiNN ($67.53$ms) and E($n$)-GNN ($56.66$ms) on the PSR dataset, while the inference on the RNA benchmark takes most time with $54.80$ms for EQGAT against $90.38$ms (PaiNN) and $76.45$ms for the E($n$)-GNN architecture.\\
The inferior execution time for the PaiNN architecture is most likely attributed to the fact that the number of scalar and vector channels is set to the same value. As the number of scalar channels is set to $128$ for all Atom3D models, the PaiNN architecture implements vector features of size $[\textsc{\#Nodes}, 3, 128]$ which becomes very memory-intensive for large graphs and requires a higher amount of FLOPS in its convolutional layers. Despite the inferior performance of the PaiNN architecture on the ATOM3D benchmark, it is worth highlighting that the PaiNN model was initially designed to predict potential energy surfaces of small molecules with high fidelity, while its architectural components most likely require further re-implementation to be usable on larger complexes. 

\section{Conclusion}
In this work, we introduce a novel attention-based graph neural network for the prediction of molecular properties of systems with varying size. Our proposed equivariant Graph Attention Network makes use of rotationally equivariant features in their intermediate layers to faithfully represent the geometry of the data, while being computationally efficient, as all equivariant functions are directly implement in the original Cartesian space. We demonstrate the potential of EQGAT by succesfully applying it to a wide range of molecular tasks, such as the prediction of quantum mechanical properties of small molecules, but also on learning problems that involve large systems such as proteins, or protein-ligand complexes. 

\section*{Code and Data Availability}
The code and data will be made available upon official publication.

\bibliography{references.bib}
\bibliographystyle{icml2022}

\newpage
\appendix
\onecolumn

\section*{Encoder Details and Hyperparameters}
All EQGAT models in this paper were trained on a single Nvidia Tesla V$100$ GPU.
\begin{center}
    \begin{table}[ht!]
    \caption{Description of architectural parameters on the QM9 and ATOM3D benchmarks.}
        \centering
        \begin{tabular}{lcccc}
             Parameter& QM9 & LBA & PSR & RSR  \\
            \toprule
             Learning rate (lr.) & $5 \cdot 10^{-4}$ & $10^{-4}$& $10^{-4}$& $10^{-4}$\\
             Maximum epochs & $300$ & $20$ & $20$&  $20$ \\
             Early stopping patience & $20$ & $10$ & $10$&  $10$ \\
             Lr. patience  & $5$ & $4$ & $4$ & $4$ \\
             Lr. decay factor & $0.75$ & $0.5$ & $0.5$ & $0.5$ \\
             Batch size  & $128$ & $16$ & $16$ & $16$ \\
             Num. layers  & $7$ & $5$ & $5$ & $5$ \\
             Num. RBFs  & $20$ & $16$ & $16$ & $16$ \\
             Cutoff [$\angstrom$] & $5.0$ & $4.5$ & $4.5$ & $4.5$ \\
             Scalar channels $F_s$  & $128$ & $128$ & $128$ & $128$ \\
             Vector channels $F_v$  & $32$ & $16$ & $16$ & $16$ \\
             \midrule
             Num. parameters & $1.1$M &  $741$k& $795$k  & $795$k 
        \end{tabular}
        
        \label{tab:my_label}
    \end{table}
\end{center}
Our model implements a Layer Normalization tailored for scalar and vector features as proposed by \cite{jing2021learning} and is applied at the beginning of every EQGAT convolutional layer.

\section{Proof Equivariance}\label{sec:appendix-rotation-equiv}
We prove the rotation equivariance for the two tensors $\{\vec{\mathbf{y}}_{0, ji}, \vec{\mathbf{y}}_{1, ji}\}$ in Equations \eqref{eq:equivariant_feature_creation-0} and \eqref{eq:equivariant_feature_creation-1} by analyzing the properties of the tensor product $\otimes$. To recap, the computation is proceed as follows
\begin{align*}
    \vec{\mathbf{y}}_{0, ji} = \vec{p}_{e, ji} \otimes \mathbf{m}_{v_{0}, ji} = \vec{p}_{e, ji} \mathbf{m}_{v_{0}, ji}^\top    ~~\in \R^{3 \times F_v},
\end{align*}
as well as
\begin{align*}
   \vec{\mathbf{y}}_{1, ji} = (\mathbf{1} \otimes \mathbf{m}_{v_1, ji}) \odot  \vec{\mathbf{a}}_{v, ji} = (\mathbf{1} \mathbf{m}_{v_1, ji}^\top) \odot  \vec{\mathbf{a}}_{v, ji}~,
\end{align*}
where $\vec{\mathbf{a}}_{v, ji} = \vec{\mathbf{v}}_{j} \times \vec{\mathbf{v}}_{i} + \vec{\mathbf{v}}_{v, j}$ is obtained via the cross-product between two vector features.\\
Since geometric tensors considered in this work are of type-$1$ only, the group action represented as rotation matrix $\mathbf{R} \in \textsc{SO}(3)$ only acts on $\vec{p}$ as well as $\vec{\mathbf{v}}$, where we excluded indices for brewity and better reading.
If the point cloud is rotated, as defined in Eq. \eqref{eq:so3-action}, (relative) position as well vector features change to
\begin{align}
    &\vec{p} \xrightarrow{G} \mathbf{R} \vec{p}~, \nonumber \\
    &\vec{\mathbf{v}} \xrightarrow{G} \mathbf{R}\vec{\mathbf{v}}~,
\end{align}
while the cross product between two vector features $\vec{\mathbf{v}}_0, \vec{\mathbf{v}}_1$ is invariant to rotation, resulting to the property
\begin{align*}
    (\mathbf{R}\vec{\mathbf{v}}_0 \times \mathbf{R}\vec{\mathbf{v}}_1) = \mathbf{R}(\vec{\mathbf{v}}_0 \times \vec{\mathbf{v}}_1)~.
\end{align*}
The message tensor $\mathbf{m}$ is an $\textsc{SO}(3)$ invariant embedding and is not affected by any rotation, but its group representation acting on $\mathbf{m}$ is defined as the identity map.\\
With definition of the tensor product $\otimes$ in Eq. \eqref{eq: tensor-product}, we can conclude that
\begin{equation}
    (\mathbf{R}\vec{p}) \otimes \mathbf{m} = (\mathbf{R}\vec{p})\mathbf{m}^\top = \mathbf{R}(\vec{p}\mathbf{m}^\top) = \mathbf{R} (\vec{p} \otimes \mathbf{m}) = \mathbf{R} \vec{\mathbf{y}}_0~, 
\end{equation}
which proves rotation equivariance for the first equation.\\
The second equation changes to the following expression when a rotation is performed:
\begin{align}
    (\mathbf{1} \otimes \mathbf{m}) \odot  (\mathbf{R}\vec{\mathbf{v}}_0 \times \mathbf{R}\vec{\mathbf{v}}_1 + \mathbf{R}\vec{\mathbf{v}}) &= (\mathbf{1} \mathbf{m}^\top) \odot \mathbf{R}(\vec{\mathbf{v}}_0 \times \vec{\mathbf{v}}_1) + \mathbf{R}\vec{\mathbf{v}} = (\mathbf{1} \mathbf{m}^\top) \odot \mathbf{R}\vec{\mathbf{a}} \\
    &= \mathbf{R} \left (\mathbf{1}\mathbf{m}^\top \odot \vec{\mathbf{a}} \right) = \mathbf{R} \left (\mathbf{1} \otimes \mathbf{m} \odot \vec{\mathbf{a}} \right) = \mathbf{R}\vec{\mathbf{y}}_1~,
\end{align}
where we use the property that the elementwise product $\odot$
between $\mathbf{1}\mathbf{m}^\top \in \R^{3\times F_v}$ and $\mathbf{R}\vec{\mathbf{a}} \in \R^{3 \times F_v}$ is applied elementwise, allowing to factorization of the rotation matrix $\mathbf{R}$.\\
As the summation of $\mathbf{R}\vec{\mathbf{y}}_0$ with $\mathbf{R}\vec{\mathbf{y}}_1$ is an equivariant function, we conclude that $\mathbf{R}(\vec{\mathbf{y}}_0 + \vec{\mathbf{y}}_1)$ is equivariant.

\section{QM9}\label{sec:appendix-qm9}
\begin{center}
    \begin{table}[ht!]
        \centering
        \caption{QM$9$ Mean Absolute Error (MAE) on the test set. Models displayed with \text{*} use different (random) train/val and test splits. For our EQGAT, we report averaged MAEs over 3 runs.}
        \begin{tabular}{lcccccccccccc}
            \hline
             Tasks & $\alpha$ & $\Delta \epsilon$ & $\epsilon_{\textsc{HOMO}}$& $\epsilon_{\textsc{LUMO}}$ & $\mu$ & $C_{\nu}$ & $G$ & $H$ & $\langle R^2 \rangle$ & $U$ & $U_0$ & ZPVE\\
             Units & bohr$^3$ & meV & meV & meV & $D$ & $\frac{\text{cal}}{\text{mol K}}$ & meV & meV & bohr$^2$ & meV & meV & meV \\
            \hline
            NMP & .092 & 69 & 43 & 38 & .030 &  .040 & 19 & 17 & .180 & 20 & 20 &  1.50 \\
            SchNet$^*$ & .235  & 63 & 41 & 34 & .033 & .033 & 14 & 14 & .073 & 19 & 14 &  1.70  \\
            Cormorant & .085 & 61 & 34 & 38 & .038 &  .026 & 20 & 21 & .961 & 21 & 22 &  2.03  \\
            LieConv & .084 & 49 & 30 & 25 & .032 &  .038 & 22 & 24 & .800 & 19 & 19 &  2.28 \\
            DimeNet++$^{\text{*}}$ & .044 & 33 & 25 & 20 & .030 &  .023 & 8 & 7 & .331 & 6 & 6 &  1.21  \\
            SE(3) Tr. & .142 & 53 & 35 & 33 & .051 & .054 & - & - & - & - & - &  -    \\
            E($n$)-GNN & .071 & 48 & 29 & 25 & .029 &  .031 & 12 & 12 & .106 & 12 & 11 &  1.55\\
            ET$^*$& .010 & 38 & 21 & 18 & .002 &  .026 & 7.64 & 6.48 & .015 & 6.30 & 6.24 & 2.12 \\
            PaiNN$^{\text{*}}$ & .045 & 46 & 28 & 20 & .012 & .024 & 7.35 & 5.98 & .066 & 5.83 & 5.85 & 1.28\\
            \hline
            EQGAT& .063 & 44 & 26 & 22 & .014 & .027 & 12 & 13 & .257 & 13 & 13 & 1.50 \\
            EQGAT$^{*}$& .053 & 32 & 20 & 16 & .011 & .024 & 23 & 24 & .382 & 25 & 25 & 2.00\\
            \hline
        \end{tabular}
        \label{tab:QM9-results-full}
    \end{table}
\end{center}

For the QM$9$ benchmark \cite{qm9}, we used the pre-processed dataset partitions from \cite{anderson2019cormorant,satorras2021en}. Following \cite{satorras2021en}, we normalized all properties by subtracting the mean and dividing by the Mean Absolute Deviation. For all targets, we only retrieve the scalar-embedding $\mathbf{s}^{(L)}_{i} \in \R^{128}$ from the last GNN layer and pool all (scalar)-node embeddings through summation to obtain the graph embedding
\begin{equation}\label{eq:graph-pooling}
\mathbf{g} = \sum_{i} \mathbf{s}_i^{(L)}~.
\end{equation}
The graph embedding $\mathbf{g}$ is consequently used to predict the target property using a 2-layer MLP following the steps as \\
$\mathbf{g} \xrightarrow{} \textsc{Linear(128, 128)} \xrightarrow{} \textsc{SiLU} \xrightarrow{} \textsc{Linear(128, 1)} \xrightarrow{} \textsc{Property}$.\\
We further note that our proposed EQGATs produce results on par with the best performing methods
on the non-energy variables, however perform inferior to the state of the art on the energy variables $(G, H, U ,
U_0)$. We hypothesize that these targets could benefit from more involved problem-tailored output-layers and additional optimization strategies, such as those compared against with.\\
As the comparison of our architecture to recent methods is not straightforward due to the usage of varying data splits, we also performed the experiments on random splits from the QM9 dataset, by creating splits with the same size as in the initial setting of \cite{anderson2019cormorant}. We noticed performance discrepancy between the EQGAT and EQGAT* model on the energy predictions, when the later model was trained on 3 random splits of the entire QM9 dataset. As we mostly refer and compare to the \cite{anderson2019cormorant} datasplits retrieved from the Github repository of the authors from \cite{satorras2021en}, we observe similar perfomance between EQGAT and E$(n)$-GNN, while ours performs better on the (norm of the dipole moment) $\mu$ target, which is in fact a vector-quantity.\\
For better comparison, we hope that in the future pre-processed sets for the training-, validation- and test splits are used.
\section{ATOM3D}\label{sec:appendix-atom3d}
We use the provided datasplits from the ATOM3D benchmark \cite{townshend2021atom3d} and their open-source python package to process the training, validation and test splits.
The LBA dataset consists on average of $382$ atoms, while the PSR and RSR datasets contains larger graphs with an average size of $1384$ and $2161$. Statistics are taken from \cite{jing2021equivariant}. \\
For the LBA task, we extract the scalar and vector embeddings, $\mathbf{x}^{(L)} = (\mathbf{s}^{(L)}, \vec{\mathbf{v}}^{(L)})$ from the GNN encoder's last layer and implement a 2-layer MLP node decoder using Gated Equivariant layers as proposed in \cite{schutt2021equivariant} that outputs scalar and vector features of dimensionality 1, i.e., ${\tilde{s}}_{i} \in \R$ and $\vec{\tilde{v}}_i \in \R^3$ for all nodes $i=1, \dots, N.$ For the final prediction $\hat{y}$ we calculate
\begin{equation}
    \hat{y} = \sum_{i=1}^N  || \tilde{s}_i \otimes \vec{p}_i + \vec{\tilde{v}}_i||_{2}^{2}~,
\end{equation}
assuming the center of mass being $\vec{p}=0$.

The decoder network for the PSR and RSR tasks are implemented using a single layer of Gated Equivariant Layer on the GNN encoder's last layer output $\mathbf{x}_{i}^{(L)}$ returning $\tilde{\mathbf{x}}_{i}^{(L)}$ which is then required for mean-pooling, where only the output scalar feature $\mathbf{s} = \frac{1}{N} \sum_{i=1}^N \mathbf{\tilde{s}}_i$ is used as input for a 2-layer MLP as follows
$\mathbf{s} \xrightarrow{} \textsc{Linear(128, 64)} \xrightarrow{} \textsc{SiLU} \xrightarrow{} \textsc{Linear(64, 1)} \xrightarrow{} \textsc{Property}$.\\
As described in the E($n$)-GNN \cite{satorras2021en} model trained on the QM$9$ dataset, we implement and train the invariant E($n$)-GNN model without the positional coordinates update equation on all three ATOM3D tasks. To enable stable training for the E($n$)-GNN model, we further transform raw distances between nodes $j$ and $i$ using a reciprocal function $e (d_{ji}) = \frac{1}{1 + d_{ji}}$ as opposed to the authors initial implementation using the squared euclidean distance.\\ 
We noticed that including layer normalization between consecutive convolutional layers for the E$(n)$-GNN and PaiNN architectures enabled stable training on the PSR and LBA benchmark datasets. This is mostly related to the fact that macromolecules consists of larger neighbourhoods when a radial cutoff is applied. Nonetheless, we still perform a sum-aggregation for the message embedding of target node $i$, such that the magnitude of neighboring embeddings is recognized by the network. 
\end{document}